\pdfoutput=1
\documentclass{article}

\usepackage[accepted]{amd}

\newcommand{\AMDlogo}{\includegraphics[height=0.28in]{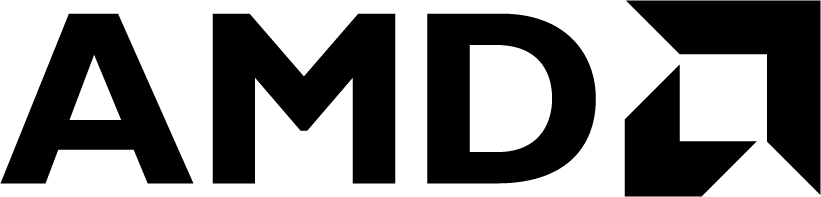}}

\usepackage[utf8]{inputenc}
\usepackage[T1]{fontenc}

\usepackage{amsmath}
\usepackage{amssymb}
\usepackage{amsfonts}
\usepackage{booktabs}
\usepackage{graphicx}
\usepackage{multirow}
\usepackage{xcolor}
\usepackage{microtype}
\usepackage{nicefrac}
\usepackage{url}
\usepackage{hyperref}
\usepackage{enumitem}
\usepackage[capitalize,noabbrev]{cleveref}
\hypersetup{colorlinks=true, linkcolor=black, citecolor=black, urlcolor=blue}

\newcommand{\ours}{PRISM-BENCH}
\newcommand{\smart}{CORE}
\newcommand{\smarttwo}{CORE-PRM}


\amdtitlerunning{PRISM-BENCH: Diagnosing Grounded and Faithful Multimodal Reasoning}

\begin{document}

\icmltitle{\ours{}: Diagnosing Grounded and Faithful\\ Multimodal Reasoning Beyond Final-Answer Accuracy}

\begin{icmlauthorlist}
\icmlauthor{Vansh Bhatia}{amd,cor}\quad
\icmlauthor{Mehdi Rezagholizadeh}{amd,cor}\quad
\icmlauthor{Parsa Ashrafi Fashi}{amd}\\[2pt]
\icmlauthor{Igor Gilitschenski}{uoft}\quad
\icmlauthor{Vikram Appia}{amd}\quad
\icmlauthor{Emad Barsoum}{amd}
\end{icmlauthorlist}

\begin{center}
$^{1}$Advanced Micro Devices, Inc. \qquad $^{2}$University of Toronto\\[3pt]
{\small \amdcorrespondingsymbol\ Corresponding authors: \texttt{\{vansh.bhatia, mehdi.rezagholizadeh\}@amd.com}}
\end{center}
\vspace{0.5\baselineskip}

\begin{abstract}
Final-answer accuracy reports whether a vision-language model was right, but not whether its answer was grounded in the image, whether its stated reasoning reflects that evidence, or what the answer cost to produce. Those are the properties that decide whether a system can be deployed. We present \ours{}, a benchmark of 1{,}000 expert-curated questions spanning five under-represented reasoning categories, in which every item carries an idealized reasoning trace written by one human annotator and validated by another. The traces support two diagnostics that accuracy cannot provide. A perception probe, in which the model answers from a description it wrote itself, attributes roughly 85\% of frontier-model failures to misperception and places the bottleneck at grounding. Trace alignment predicts answer correctness for all 25 systems tested (macro AUROC 0.66) and survives length control, so it measures reasoning content and not verbosity. We further introduce \smart{}, an efficiency-aware metric that reorders 24.2\% of model pairs relative to accuracy while remaining uncorrelated with response length. Across 25 open and closed systems, including GPT-5 and Claude Opus 4.6, the best accuracy is 66.0\%. On the four categories with a human baseline the best system trails humans by 18.4 points, and recent frontier gains fall on recognition and not on spatial grounding.

\smallskip
\textbf{Dataset:} \url{https://huggingface.co/datasets/vanshbhatia/PRISM-BENCH}
\end{abstract}

\section{Introduction}
\label{sec:intro}

Vision-language models are moving from describing a scene to acting within it, in robotics, driver assistance, and clinical decision support. That shift changes what an evaluation has to report. Correctness remains necessary, but a deployed system must also be \textbf{grounded}, so that its answer follows from what the image contains and not from a linguistic prior; \textbf{faithful}, so that its stated reasoning reflects the evidence it actually used; and \textbf{affordable}, now that reasoning-heavy inference dominates serving cost.

Standard multimodal benchmarks report none of these properties, because they score only the final answer. A model that arrives at the right option through a hallucinated causal story is indistinguishable from one that reasons soundly, and a model that spends ten thousand tokens looks identical to one that spends a hundred. In deployment these are different systems. The first fails silently once the input distribution shifts; the second fails the latency budget.

\begin{figure}[t]
\centering
\includegraphics[width=\textwidth]{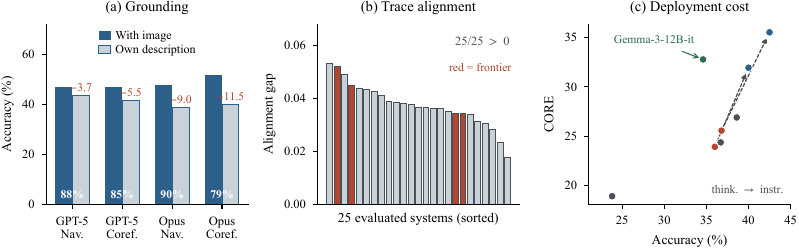}
\caption{Three deployment-relevant diagnostics that final-answer accuracy cannot provide. \textbf{(a)}~Accuracy \emph{falls} when a model must answer from its own image description instead of the image (drop annotated in red); the white figure inside each bar is the share of that model's baseline failures that the model itself, judging in a fresh context, attributes to perception instead of inference. \textbf{(b)}~For all 25 evaluated systems, responses that are correct align more closely with the human reasoning trace than responses that are incorrect (macro AUROC 0.66, and 0.665 after regressing out response length). \textbf{(c)}~\smart{} reorders models relative to accuracy; dashed arrows connect matched thinking/instruct pairs, which spend more tokens for less accuracy.}
\label{fig:diagnostics}
\end{figure}

We present \ours{}, a benchmark built so that these properties can be measured. It contains 1{,}000 expert-curated questions across five under-represented reasoning categories, and each item is paired with an idealized reasoning trace written by a human annotator: the minimal, logically complete chain a careful person would follow. Because that reference supplies a per-item ground truth for the reasoning process and not only for the outcome, it makes diagnostics available that an accuracy-only benchmark cannot support.

\Cref{fig:diagnostics} summarizes the three diagnostics this makes possible. Our contributions are:
\begin{itemize}[leftmargin=*,itemsep=1pt]
\item \textbf{The benchmark.} 1{,}000 questions across five categories, each with a ground-truth answer, a human-authored reasoning trace, and multi-label reasoning tags, built by a four-stage pipeline in which no annotator validates an item they authored.
\item \textbf{A grounding diagnostic.} A perception probe attributes roughly 85\% of frontier failures to misperception, and accuracy falls by up to 11.5 points when a model must work from its own description of the image.
\item \textbf{A reasoning-content diagnostic.} Reference-trace alignment separates correct from incorrect responses for all 25 systems tested, at macro AUROC 0.66 and 0.665 after length control.
\item \textbf{A cost-aware metric.} \smart{} scores accuracy per unit of reasoning cost, reordering 24.2\% of model pairs relative to accuracy while remaining uncorrelated with raw length ($\rho = -0.05$).
\item \textbf{Evaluation and self-audit.} 25 open and closed systems, including GPT-5 and Claude Opus 4.6, against an expanded human baseline, with a reported audit of our protocol's ranking stability, critic reliability, and contamination exposure.
\end{itemize}

\section{Related Work}
\label{sec:related}

\paragraph{Multimodal reasoning benchmarks.} A rich ecosystem probes visual mathematics~\citep{lu2024mathvista,zhang2024mathverse,qiao2024we}, spatial and relational understanding~\citep{omnispatial25,mm-spatial}, abstraction~\citep{cai2025mm,li2024omnibench}, and domain-specific settings~\citep{luo2025finmme,meng2025mm,bi2025verify,hao2025can}. Broader suites such as MMMU-Pro~\citep{yue2024mmmu} and MegaBench~\citep{chen2025mega-bench} reduce shortcut exploitation. These benchmarks largely score final answers, rarely release step-level reference reasoning, and do not report generation cost. As a result they cannot separate a grounded solution from a lucky one, which is the distinction that matters at deployment.

\paragraph{Reasoning and process supervision.} Chain-of-thought prompting~\citep{zhang2024multimodalchainofthoughtreasoninglanguage} and RL-style reasoning training~\citep{deepseekai2025deepseekr1incentivizingreasoningcapability,muennighoff2025s1simpletesttimescaling} have been extended to multimodal models via GRPO-style optimization~\citep{tan2025reasonrftreinforcementfinetuningvisual,wan2025srpoenhancingmultimodalllm}, self-reflection~\citep{wang2025vlrethinkerincentivizingselfreflectionvisionlanguage}, and structured captioning~\citep{xia2025visionaryr1mitigatingshortcutsvisual}. Process reward models including VisualPRM~\citep{wang2025visualprmeffectiveprocessreward} and DreamPRM~\citep{cao2025dreamprm15unlockingpotentialinstance} score intermediate steps, but current PRMs show unstable labelling and weak out-of-distribution generalization~\citep{sun2025mmverifyenhancingmultimodalreasoning}, a limitation we quantify on our own data in \Cref{sec:reliability}. Test-time compute scaling~\citep{snell2024scaling} makes cost-aware evaluation increasingly necessary, and verbose reasoning has been linked to hallucination~\citep{sun2024verbosity}.

\section{\ours{}}
\label{sec:benchmark}

\subsection{Categories}

\begin{figure}[t]
\centering
\includegraphics[width=0.66\textwidth]{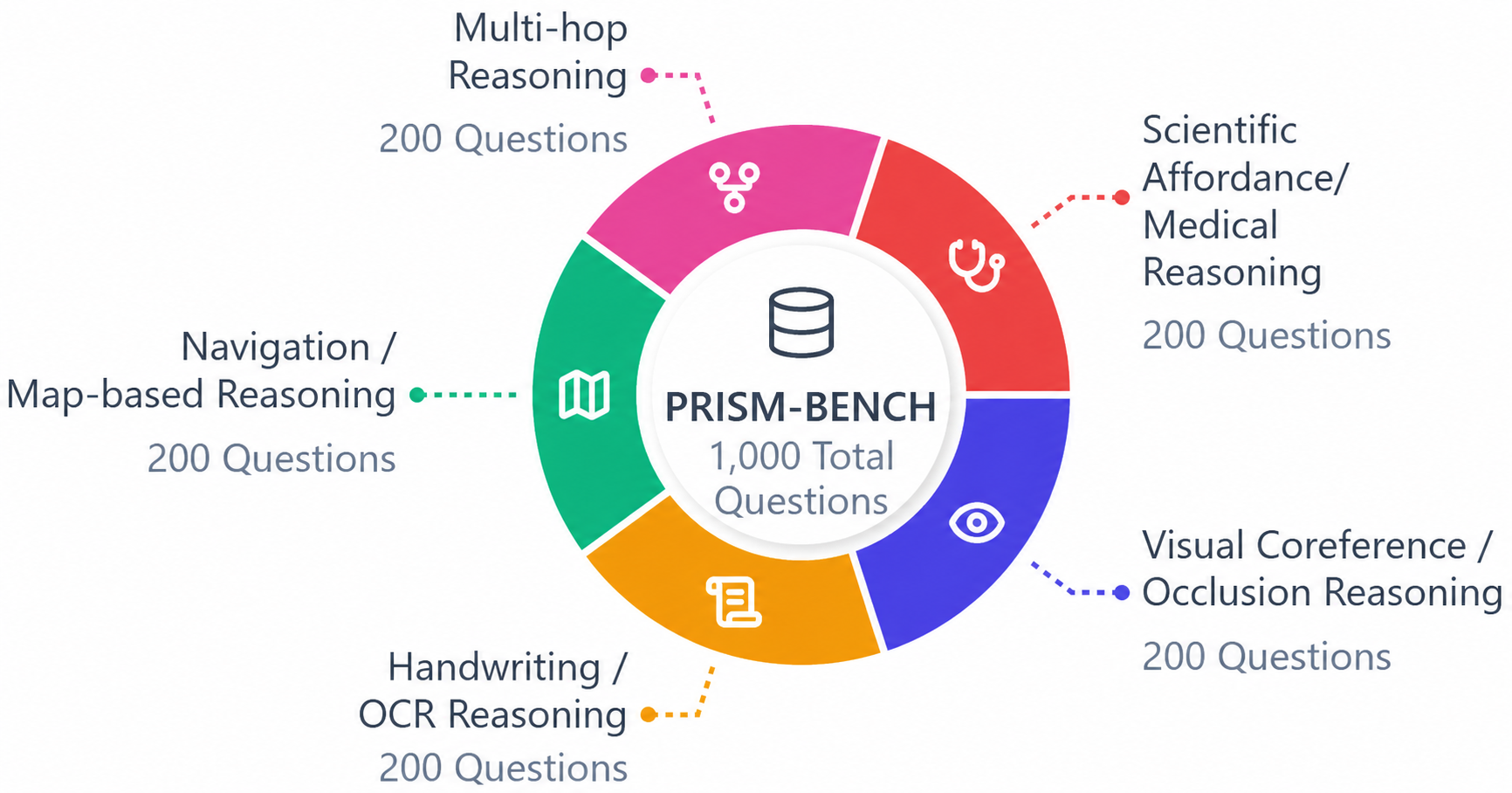}
\caption{The five reasoning categories, balanced at 200 items each. The categories mix functional reasoning patterns, such as multi-hop composition, with domain-specific grounding requirements, such as medical interpretation, because deployment demands both.}
\label{fig:categories}
\end{figure}

\ours{} contains 1{,}000 manually curated questions, 200 in each of five categories (\Cref{fig:categories}), chosen to probe what we call the \emph{Perception--Cognition Disconnect}: cases where a model perceives visual elements yet fails to build a coherent world model. We deliberately exclude saturated categories such as chart and math QA, where memorized templates are exploitable. \Cref{tab:datastats} gives the dataset statistics.

\begin{table}[t]
\centering
\footnotesize
\caption{\ours{} dataset statistics. The three answer-choice rows partition all 1{,}000 items; most are 4-way or wider multiple choice, and 11 items are open-ended or binary.}
\label{tab:datastats}
\begin{tabular}{ll}
\toprule
\textbf{Statistic} & \textbf{Value} \\
\midrule
Total questions & 1{,}000 \\
Categories & 5 (200 each) \\
Average question length & 24.7 words \\
Average trace length (structured / free-form) & 70.1 / 43.8 words \\
Questions with 0--2 answer choices & 11 \\
Questions with 3--4 answer choices & 245 \\
Questions with 4+ answer choices & 744 \\
Annotation time per question & 20--30 min \\
Image sources & RefCOCO, OpenStreetMap, MedXpertQA, \\
 & handwriting volunteers, IAM \\
\bottomrule
\end{tabular}
\end{table}

\begin{enumerate}[leftmargin=*,itemsep=1pt]
\item \textbf{Visual Coreference / Occlusion} (200): resolving references to partially occluded targets among distractors, requiring object identity and spatial logic under incomplete evidence.
\item \textbf{Handwriting / OCR Reasoning} (200): recognizing handwritten text and reasoning over it, where one misread character cascades downstream.
\item \textbf{Navigation / Map-based Reasoning} (200): interpreting schematic maps and composing multi-step spatial instructions.
\item \textbf{Multi-hop Reasoning} (200): chaining visual and textual clues across image regions.
\item \textbf{Scientific / Medical Reasoning} (200): interpreting medical visual cues with domain causal knowledge, where models often perceive correctly but hallucinate causal explanations.
\end{enumerate}

Each item carries optional answer choices, a ground-truth answer, a human-authored idealized reasoning trace, and multi-label reasoning tags (visual, spatial, textual, commonsense). The trace is recorded in two human-written forms: the annotator's original free-form chain (mean 43.8 words) and a structured step-by-step rendering of the same chain (mean 70.1 words). \smart{} takes its length reference from the structured form, and trace alignment compares against the original.

\subsection{Construction}

Images come from RefCOCO~\citep{refcoco} (coreference, multi-hop), schematic layouts derived from OpenStreetMap (navigation), handwriting scans and IAM, and MedXpertQA~\citep{medxpertqa} (medical). A four-stage pipeline enforces separation of duties, so no annotator both authors and validates the same item: (1)~experts design questions that decouple surface cues from the reasoning target; (2)~annotators author idealized minimal traces, manually audited and edited by experts; (3)~annotators assign multi-label tags; (4)~a \emph{separate} annotator validates answer correctness, trace validity, and clarity, with failing items revised or removed. Each item requires 20--30 minutes of total annotation effort.

Question, option, answer, and trace text is newly authored for 807 of 1{,}000 items (80.7\%); only source images are reused. The 193 medical items reuse an expert-validated image, question, and answer, with the trace authored to that verified answer.

\section{Cost-Aware Metrics}
\label{sec:metrics}

Two models with equal accuracy can differ by an order of magnitude in tokens. \smart{} scores correctness per unit reasoning cost, using the human trace length as the per-item reference:
\begin{equation}
\smart{}(a,\mathbf{l}) = \frac{a}{\frac{1}{N}\sum_{i=1}^{N} \frac{\log l_i}{\log L_i^{\mathrm{gt}}}},
\label{eq:core}
\end{equation}
where $a$ is accuracy, $l_i$ the generated length, and $L_i^{\mathrm{gt}}$ the human trace length. When $l_i = L_i^{\mathrm{gt}}$ the denominator is 1 and \smart{} reduces to accuracy. We normalize in log space because length distributions are heavy-tailed, so that a single verbose response cannot dominate the average. The reference serves as an efficiency floor and is not a claim that the human path is the only valid one: the log ratio tolerates alternative correct derivations of moderately different length.

The critic-augmented variant incorporates step-level process reward scores $v_i \in [0,1]$ to discount correct-but-ungrounded answers:
\begin{equation}
\smarttwo{}(\mathbf{a},\mathbf{v},\mathbf{l}) = \frac{1}{2N}\sum_{i=1}^{N} \frac{(a_i+v_i)\log L_i^{\mathrm{gt}}}{\log l_i}.
\label{eq:coreprm}
\end{equation}
\smart{} is critic-free and is our primary metric; \smarttwo{} is a critic-agnostic template whose validity is bounded by the judge supplied to it (\Cref{sec:reliability}).

\section{Experiments}
\label{sec:experiments}

Models generate an answer plus a step-by-step rationale, via chain-of-thought prompting or native thinking mode. Open-weight models use greedy decoding, giving zero run-to-run variance, and all models share a fixed prompt template and answer-extraction cascade.

\subsection{Frontier Models Remain Far Below Humans}

\begin{table}[t]
\centering
\footnotesize
\caption{Accuracy (\%) of all 25 evaluated systems across the five \ours{} categories, against the human baseline (5 annotators $\times$ 50 items per category). No system exceeds 66\% overall, or 84\% in any single category. The newest frontier systems lead on OCR yet remain near 47\% on navigation, where humans reach 86\%: recent gains concentrate on recognition rather than spatial grounding.}
\label{tab:fullresults}
\begin{tabular}{llcccccc}
\toprule
\textbf{Model} & \textbf{Size} & \textbf{Overall} & \textbf{Coref.} & \textbf{OCR} & \textbf{Nav.} & \textbf{Multi-hop} & \textbf{Sci./Med.} \\
\midrule
\multicolumn{8}{l}{\textit{Closed-source}} \\
Gemini 2.5 Pro & -- & \textbf{66.0} & \textbf{57.5} & 77.0 & \textbf{58.2} & 69.2 & \textbf{68.5} \\
GPT-5 & -- & 63.6 & 47.0 & 82.0 & 47.0 & \textbf{73.6} & 68.0 \\
Claude Opus 4.6 & -- & 62.4 & 51.5 & \textbf{83.5} & 47.8 & 69.7 & 59.3 \\
OpenAI o3 & -- & 57.2 & 39.5 & 75.5 & 46.8 & 65.2 & 59.5 \\
Claude Sonnet 4 & -- & 39.6 & 39.0 & 46.5 & 39.8 & 38.3 & 34.5 \\
\multicolumn{8}{l}{\textit{Open-weight (instruct)}} \\
Qwen3-VL-30B-A3B-Inst. & 30B & 42.5 & 41.0 & 52.5 & 35.3 & 48.3 & 35.5 \\
Qwen3-VL-8B-Inst. & 8B & 40.0 & 39.5 & 50.5 & 42.3 & 40.3 & 27.5 \\
Gemma-3-12B-it & 12B & 34.6 & 39.0 & 40.5 & 27.4 & 38.3 & 28.0 \\
InternVL3.5-8B & 8B & 30.2 & 29.5 & 36.5 & 27.4 & 30.9 & 27.0 \\
Gemma-3-4B-it & 4B & 29.0 & 36.0 & 30.5 & 23.9 & 32.8 & 22.0 \\
Qwen2.5-VL-7B-Inst. & 7B & 27.4 & 30.0 & 31.5 & 32.8 & 22.9 & 20.0 \\
InternVL3-8B-Inst. & 8B & 26.7 & 30.5 & 25.0 & 26.9 & 30.9 & 20.5 \\
Qwen2.5-VL-3B-Inst. & 3B & 22.1 & 24.5 & 19.0 & 28.9 & 19.4 & 18.5 \\
InternVL3-2B-Inst. & 2B & 20.7 & 31.0 & 16.0 & 22.4 & 23.9 & 10.0 \\
Molmo-7B-D-0924 & 7B & 20.1 & 25.5 & 21.0 & 20.9 & 19.4 & 14.0 \\
\multicolumn{8}{l}{\textit{Open-weight (thinking / RL-trained)}} \\
Skywork-R1V3-38B & 38B & 38.6 & 42.5 & 42.5 & 32.3 & 44.3 & 31.5 \\
Qwen3-VL-30B-A3B-Think. & 30B & 36.8 & 34.0 & 57.5 & 36.8 & 38.8 & 17.0 \\
Skywork-R1V2-38B & 38B & 36.7 & 40.0 & 32.5 & 30.9 & 48.8 & 31.5 \\
Ovis2.5-9B & 9B & 36.1 & 36.5 & 40.5 & 37.8 & 42.8 & 22.5 \\
Qwen3-VL-8B-Think. & 8B & 36.0 & 32.0 & 54.0 & 33.3 & 33.3 & 27.5 \\
VL-Rethinker & 7B & 29.1 & 32.0 & 35.0 & 31.3 & 30.4 & 16.5 \\
Kimi-VL-A3B-Think. & 16B & 29.0 & 34.5 & 33.5 & 32.8 & 26.9 & 17.5 \\
Ovis2.5-2B & 2B & 28.4 & 32.5 & 30.5 & 32.3 & 29.4 & 17.0 \\
Visionary-R1 & 7B & 26.2 & 30.5 & 23.0 & 31.8 & 24.4 & 21.5 \\
GLM-4.1V-9B-Think. & 9B & 23.8 & 29.0 & 18.5 & 24.4 & 27.4 & 19.5 \\
\midrule
\textit{Human baseline} & -- & \textit{83.9} & \textit{84.0} & \textit{91.2} & \textit{86.0} & \textit{74.4} & n/a \\
\bottomrule
\end{tabular}
\end{table}

\begin{figure}[t]
\centering
\includegraphics[width=\textwidth]{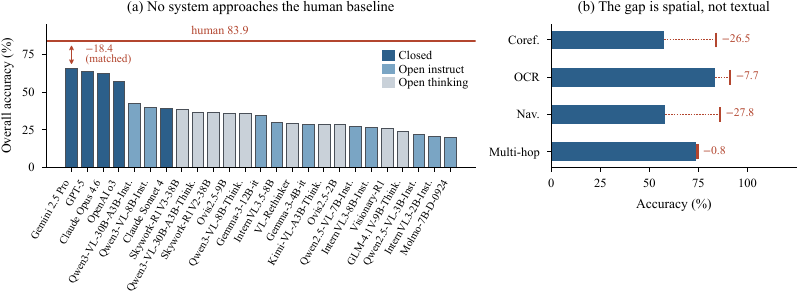}
\caption{\textbf{(a)}~Overall accuracy for all 25 evaluated systems, shaded by family, against the 83.9\% human baseline (red). Restricted to the four categories where both are measured, the best system trails humans by 18.4 points, and the open/closed gap is larger than any difference within the open families. \textbf{(b)}~Best system of any kind versus the human baseline (red tick) per category. The deficit is concentrated in the two spatially grounded categories and nearly vanishes on multi-hop, where the reasoning is largely textual once the scene is read.}
\label{fig:humangap}
\end{figure}

No model exceeds 66\% overall, against a human average of 83.9\% (\Cref{tab:fullresults}). The two aggregates cover different category sets, since the human study excludes Scientific/Medical, so the like-for-like comparison is over the four categories both measure: there the best system reaches 65.5\%, a gap of 18.4 points. The shape of that gap matters more than its size. The newest frontier systems post the strongest OCR scores, above Gemini's, yet their navigation scores are the weakest of the four, near 47\% where humans reach 86\%. Recent progress has concentrated in recognition, while spatial grounding, the capability most relevant to embodied and autonomous settings, has not moved.

\Cref{fig:humangap} makes this explicit. Across all 25 systems, none approaches the human baseline, and the remaining deficit is almost entirely spatial. On multi-hop, where the reasoning is largely textual once the scene has been read, the best system comes within 0.8 points of humans; on coreference and navigation it trails by 26.5 and 27.8 points. A system can look competitive on an aggregate multimodal leaderboard and still be unusable for the embodied tasks that motivate deployment. Instruct variants also match or exceed their thinking counterparts at matched scale (40.0\% against 36.0\% for Qwen3-VL-8B, and 42.5\% against 36.8\% at 30B).

Two further patterns are stable across \Cref{tab:fullresults}. Scale is not decisive: Ovis2.5-9B outperforms several 30B+ systems on navigation, so category specialization matters more than parameter count. And navigation and scientific/medical remain the hardest categories for nearly every system, while OCR is the most tractable.

\subsection{Grounding: Most Failures Are Perceptual}
\label{sec:perception}

To locate where models break, we run a two-stage probe on the two categories that depend most on grounding. The model first describes the image; it is shown the question but told not to answer it, so the description is directed at what the item needs. It then answers from that description alone, with the visual input removed. Finally, in a fresh text-only context, the same model labels each baseline failure as either perceptual or inferential (\Cref{app:probe}).

\begin{table}[t]
\centering
\small
\caption{Perception-vs-reasoning probe on navigation and coreference. Accuracy \emph{falls} when models answer from their own descriptions, and the large majority of failures are attributed to perception.}
\label{tab:probe}
\begin{tabular}{llccc}
\toprule
\textbf{Model} & \textbf{Category} & \textbf{Baseline acc.} & \textbf{From own description} & \textbf{\% failures perceptual} \\
\midrule
GPT-5           & Navigation  & 47.0 & 43.3 & \textbf{88} \\
GPT-5           & Coreference & 47.0 & 41.5 & 85 \\
Claude Opus 4.6 & Navigation  & 47.8 & 38.8 & \textbf{90} \\
Claude Opus 4.6 & Coreference & 51.5 & 40.0 & 79 \\
\midrule
\textit{Combined} & Nav.\ + Coref. & ${\sim}48$--$50$ & ${\sim}39$--$42$ & $\mathbf{{\sim}85}$ \\
\bottomrule
\end{tabular}
\end{table}

Two independent signals converge (\Cref{tab:probe}, plotted in \Cref{fig:diagnostics}a). The judge attributes about 85\% of failures to perception. Independently of any judge, accuracy drops when a model must work from its own description, by 3.7 points for GPT-5 on navigation and by 11.5 for Opus on coreference, which means the descriptions omit information the model needed. Were the failures inferential, re-reading a faithful self-description would cost nothing.

The practical consequence is that adding reasoning tokens to a model that has already mis-seen the scene cannot recover the answer. Effort is better spent on visual grounding than on longer chains, which is consistent with our finding that thinking variants underperform instruct variants at matched scale.

\subsection{Reference-Trace Alignment Predicts Correctness}
\label{sec:alignment}

Does a model's stated reasoning track the evidence, or is it a narrative attached to a guess? We segment each gold trace and model response into steps, embed them, and for each gold step take the maximum cosine similarity to any model step, yielding a semantic overlap score. This uses the traces for \emph{content}, not length.

\begin{table}[t]
\centering
\small
\caption{Trace alignment predicts answer correctness. AUROC is computed per model over ${\sim}1{,}000$ items; the length-controlled column regresses out response length. All 25 systems exceed chance in both columns.}
\label{tab:align}
\begin{tabular}{lcc}
\toprule
\textbf{Model} & \textbf{AUROC (align $\rightarrow$ correct)} & \textbf{AUROC (length-controlled)} \\
\midrule
OpenAI o3                & 0.71 & 0.71 \\
GPT-5                    & 0.69 & 0.69 \\
Claude Opus 4.6          & 0.68 & 0.67 \\
Qwen3-VL-8B-Instruct     & 0.67 & 0.68 \\
Qwen3-VL-8B-Thinking     & 0.67 & 0.68 \\
Skywork-R1V3-38B         & 0.63 & 0.64 \\
\midrule
\textbf{Macro-avg (25 systems)} & \textbf{0.66} & \textbf{0.665} \\
\bottomrule
\end{tabular}
\end{table}

For all 25 systems, correct responses align more closely with the human trace than incorrect ones (\Cref{tab:align}; \Cref{fig:diagnostics}b shows the full per-model spread). The macro AUROC of 0.66 is essentially unchanged at 0.665 once response length is regressed out, and all 25 remain above chance, so the effect belongs to reasoning content and not to verbosity. The result is an association: responses that cover the reference reasoning are more often correct, which does not by itself show that the stated reasoning is what produced the answer. Alignment is highest for OCR and lowest for multi-hop, which matches the intuition that compositional reasoning admits more divergent valid routes.

This gives a critic-free signal of reasoning quality that requires no process reward model, and it doubles as a validation of the reference traces themselves: a single idiosyncratic annotator path could not predict correctness across 25 heterogeneous systems.

\paragraph{Toward hallucination detection at deployment time.} Alignment is computed from the response alone against a stored reference, so it can be evaluated without knowing the answer, which is the situation a deployed monitor is in. A response whose steps diverge from the reference is measurably more likely to be wrong, making alignment usable as a cheap abstention or escalation signal. The signal is modest, and its limits are worth stating. An AUROC of 0.66 supports flagging low-confidence outputs for review, not verifying individual items, and the measure needs a reference trace, so it applies to curated evaluation sets and not to open-world inputs. It nonetheless offers something a process reward model does not. In \Cref{sec:reliability} we show that our off-the-shelf critic is close to non-discriminative on this data, whereas trace alignment separates correct from incorrect responses for every model we tested.

\subsection{Deployment Cost: Longer Reasoning Does Not Pay}
\label{sec:core}

\begin{table}[t]
\centering
\scriptsize
\setlength{\tabcolsep}{3pt}
\caption{Cost-aware metrics for the 20 open-weight models with exact token counts. ``Norm.\ len.'' is mean response length divided by the human reference length. Ranks are computed within this table; 46 of the 190 model pairs (24.2\%) are ordered differently by accuracy and \smart{}. \smart{} is unavailable for closed systems that hide thinking tokens.}
\label{tab:corefull}
\begin{tabular}{llccccccc}
\toprule
\textbf{Model} & \textbf{Size} & \textbf{Acc.\ (\%)} & \textbf{Norm.\ len.} & \textbf{\smart{}} & \textbf{VisualPRM $v$} & \textbf{\smarttwo{}} & \textbf{Acc.\ rank} & \textbf{\smart{} rank} \\
\midrule
Gemma-3-4B-it & 4B & 29.0 & 1.04 & 28.01 & 0.64 & 45.48 & 11 & 5 \\
Gemma-3-12B-it & 12B & 34.6 & 1.06 & 32.76 & 0.69 & 49.58 & 8 & 2 \\
GLM-4.1V-9B-Think. & 9B & 23.8 & 1.26 & 18.93 & 0.26 & 20.05 & 17 & 20 \\
InternVL3-2B-Inst. & 2B & 20.7 & 1.04 & 19.83 & 0.57 & 37.60 & 19 & 18 \\
InternVL3-8B-Inst. & 8B & 26.7 & 1.06 & 25.18 & 0.65 & 43.80 & 15 & 12 \\
InternVL3.5-8B & 8B & 30.2 & 1.05 & 28.76 & 0.66 & 45.90 & 9 & 4 \\
Kimi-VL-A3B-Think. & 16B & 29.0 & 1.34 & 21.61 & 0.70 & 37.76 & 12 & 17 \\
Molmo-7B-D-0924 & 7B & 20.1 & 1.04 & 19.38 & 0.56 & 36.91 & 20 & 19 \\
Ovis2.5-2B & 2B & 28.4 & 1.26 & 22.62 & 0.55 & 34.33 & 13 & 16 \\
Ovis2.5-9B & 9B & 36.1 & 1.31 & 27.61 & 0.59 & 37.27 & 6 & 6 \\
Qwen2.5-VL-3B-Inst. & 3B & 22.1 & 0.97 & 22.85 & 0.55 & 40.68 & 18 & 15 \\
Qwen2.5-VL-7B-Inst. & 7B & 27.4 & 1.04 & 26.32 & 0.62 & 43.11 & 14 & 10 \\
Qwen3-VL-8B-Think. & 8B & 36.0 & 1.51 & 23.92 & 0.70 & 36.40 & 7 & 14 \\
Qwen3-VL-8B-Inst. & 8B & 40.0 & 1.25 & 31.92 & 0.72 & 46.26 & 2 & 3 \\
Qwen3-VL-30B-A3B-Think. & 30B & 36.8 & 1.44 & 25.56 & 0.72 & 38.52 & 4 & 11 \\
Qwen3-VL-30B-A3B-Inst. & 30B & 42.5 & 1.20 & 35.50 & 0.67 & 46.95 & 1 & 1 \\
Skywork-R1V2-38B & 38B & 36.7 & 1.51 & 24.37 & 0.69 & 35.84 & 5 & 13 \\
Skywork-R1V3-38B & 38B & 38.6 & 1.44 & 26.89 & 0.71 & 38.58 & 3 & 8 \\
Visionary-R1 & 7B & 26.2 & 0.99 & 26.52 & 0.56 & 42.15 & 16 & 9 \\
VL-Rethinker & 7B & 29.1 & 1.06 & 27.54 & 0.63 & 44.00 & 10 & 7 \\
\bottomrule
\end{tabular}
\end{table}

\Cref{tab:corefull} gives the cost-aware view for the 20 open-weight models where we can count generated tokens exactly. Closed systems are excluded because their thinking tokens are not exposed, so their reasoning cost cannot be measured; this is a limitation of provider APIs, not of the metric. The table shows the deployment picture accuracy hides. At matched scale, Qwen3-VL-8B-Thinking generates $2.5\times$ as many tokens as its instruct counterpart (1{,}812 versus 737) for \emph{lower} accuracy, and \smart{} separates them by 8 points; the same holds at 30B. Gemma-3-12B-it, ranked 8th of 20 by accuracy, ranks 2nd by \smart{} on the strength of disciplined reasoning.

These reorderings are not a repackaging of length (\Cref{fig:diagnostics}c). Across the 20 open-weight models with exact token counts, the Spearman correlation between \smart{} and normalized length is $\rho = -0.050$ ($p = 0.834$), so \smart{} is not an inverse-length score. Against accuracy it is $\rho = +0.674$ ($p = 0.001$), high enough that correctness dominates but well short of redundancy. Accuracy and normalized length are themselves correlated ($\rho = +0.657$, $p = 0.002$), but this is a confound rather than a mechanism: more capable families happen to generate longer outputs, while within a matched-scale family the thinking variant spends 2.2--2.5$\times$ as many tokens for equal or lower accuracy. \smarttwo{} tracks accuracy only weakly ($\rho = +0.313$, $p = 0.179$). In all, 24.2\% of model pairs (46 of 190) are ranked discordantly by the two measures. The mechanism is the per-item normalization in \Cref{eq:core}: verbosity on an item a human solved briefly is penalized more heavily than the same verbosity on a deep item, and any aggregate length statistic discards that distinction.

\begin{table}[t]
\centering
\footnotesize
\caption{Reflective-token analysis for 12 representative models. ``Reflective tokens'' counts hesitation and backtracking phrases (``Wait'', ``Actually'', ``Let me reconsider'') per response. More self-correction does not yield higher accuracy: Qwen3-VL-8B-Thinking averages 17.0 reflective phrases and 1{,}812 tokens to reach 36.0\%, while its instruct counterpart averages 2.6 phrases and 737 tokens to reach 40.0\%.}
\label{tab:reflective}
\begin{tabular}{lccc}
\toprule
\textbf{Model} & \textbf{Acc.\ (\%)} & \textbf{Avg.\ tokens} & \textbf{Reflective tokens} \\
\midrule
Gemma-3-4B-it & 29.0 & 195.5 & 0.1 \\
Gemma-3-12B-it & 34.6 & 198.4 & 0.1 \\
Qwen3-VL-8B-Inst. & 40.0 & 736.7 & 2.6 \\
Qwen3-VL-8B-Think. & 36.0 & 1811.7 & 17.0 \\
Qwen3-VL-30B-A3B-Inst. & 42.5 & 581.5 & 1.6 \\
Qwen3-VL-30B-A3B-Think. & 36.8 & 1253.2 & 12.0 \\
Ovis2.5-2B & 28.4 & 557.0 & 9.2 \\
Ovis2.5-9B & 36.1 & 658.6 & 11.9 \\
Skywork-R1V2-38B & 36.7 & 1380.9 & 9.8 \\
Skywork-R1V3-38B & 38.6 & 928.7 & 7.9 \\
Kimi-VL-A3B-Think. & 29.0 & 715.0 & 8.2 \\
GLM-4.1V-9B-Think. & 23.8 & 470.7 & 12.1 \\
\bottomrule
\end{tabular}
\end{table}

\paragraph{Reflection tracks verbosity, not quality.}
\label{sec:reflective}
\Cref{tab:reflective} counts hesitation and backtracking phrases per response for 12 representative models, and reinforces the cost argument from a different direction. Gemma-3 models reach 29.0--34.6\% with roughly 200 tokens and essentially no reflective phrases, while GLM-4.1V-9B-Thinking averages one self-correction every 39 tokens and posts the lowest accuracy in the table. For a deployed system this matters because reflective loops are pure cost: they lengthen the response, raise latency, and do not buy correctness.

\subsection{Reliability of Our Own Protocol}
\label{sec:reliability}

Deployment claims should state their own uncertainty, so we report two audits of the evaluation itself.

\paragraph{Ranking stability.} We evaluated three frontier systems under four benign prompt and decoding variants on a 150-item stratified subset, giving 1{,}799 runs with no errors. Each model's accuracy is stable, deviating at most 4.8 points from its own mean, but the top-1 ranking flips. GPT-5 and Gemini sit within about two points of each other and exchange first place depending on the prompt, while Claude is last under every condition. Adjacent leaderboard positions among closely matched systems are therefore not identified by accuracy alone, which is a further reason to report grounding, faithfulness, and cost beside it.

\paragraph{Critic reliability.} We tested whether VisualPRM-8B, our default critic, separates correct from incorrect reasoning on 200 balanced traces drawn from 5 categories and 8 models. It does so only weakly, with mean $v_i$ of 0.608 against 0.596, and the gap is negative in multi-hop and coreference/occlusion. \smarttwo{} should therefore be read as a critic-agnostic template and not as a settled score. This is why the critic-free \smart{} is our primary metric and carries every headline conclusion above. A stronger multimodal critic would improve \smarttwo{} without any change to the benchmark.

\paragraph{Contamination.} Because the source images are public, we ran a behavioral check: if exposure leaked answers, the most-reused sources would be the easiest. The opposite holds. GPT-5 and Opus 4.6 score 60.6 and 61.3 on COCO-RefCOCO images and 46.8 and 47.8 on OpenStreetMap schematics, making the two most public sources the two hardest, while handwriting scans, the least likely to have been memorized, are easiest at 82.1 and 83.6.

\section{Failure Modes}
\label{sec:failure}

Each category produces a distinct failure signature. In visual coreference, models over-attend to the most salient object and mishandle relative spatial phrases. In handwriting and OCR, a single misrecognized character cascades through the downstream arithmetic and corrupts an otherwise sound chain. In navigation, models identify landmarks correctly but fail at path composition, producing off-by-one directional errors and confusing left with right at intersections. In multi-hop, the first hop usually succeeds while the second introduces fabricated constraints. In the medical category, models frequently perceive the visual feature correctly and then hallucinate a causal attribution. That category has the highest rate of confidently wrong explanations and is the clearest safety concern for high-stakes deployment.

\section{Conclusion}
\label{sec:conclusion}

\ours{} evaluates multimodal reasoning along the axes that determine whether a system can be deployed: grounding, faithfulness, and cost. Pairing every question with a human-authored reasoning trace makes two diagnostics possible that accuracy cannot support. A perception probe attributes most frontier failures to grounding instead of inference, and a length-controlled alignment measure predicts correctness across 25 systems. Read together with a cost-aware metric that reorders a quarter of all model pairs, these results show that frontier systems remain far below human performance, that recent gains have concentrated in recognition while spatial grounding has stalled, and that additional reasoning tokens do not repair a scene the model has already mis-seen. We release the benchmark, traces, tags, and evaluation code at \url{https://huggingface.co/datasets/vanshbhatia/PRISM-BENCH}.

\section*{Acknowledgments and Disclosure of Funding}
This work was carried out at AMD. No external funding was received.

\bibliographystyle{plainnat}
\bibliography{refs}

\appendix

\section{Perception--Reasoning Probe Protocol}
\label{app:probe}

The probe in \Cref{sec:perception} runs in two stages on the two most grounding-dependent categories, navigation and coreference.

\begin{enumerate}[leftmargin=*,itemsep=1pt]
\item \textbf{Baseline.} The model answers the question with the image present, under the standard prompt and answer-extraction cascade used everywhere else in the paper.
\item \textbf{Describe.} The model is shown the same image together with the question and asked for an exhaustive description of everything in the image relevant to it: objects and their absolute and relative positions, visible text, labels and street names, colors, orientations, and counts. It is explicitly instructed not to answer. Because the description is conditioned on the question, it is directed at what the item needs rather than being a generic caption.
\item \textbf{Answer from description.} A fresh context receives the question and the model's own description, with the image removed, and must answer from text alone.
\item \textbf{Failure attribution.} For each item the model failed at baseline, the same model, in a fresh text-only context, is shown the question and options, the gold answer, the gold free-form reasoning trace, and its own description. It labels the primary cause of failure as \emph{perceptual} (the description omitted or misstated a visual fact needed to answer) or \emph{inferential} (the description contained the needed facts but the answer was still wrong).
\end{enumerate}

The two stages are deliberately redundant, because each has a weakness the other does not. The attribution step depends on a judge, so it inherits whatever bias that judge has. The accuracy comparison between stages 1 and 3 depends on no judge at all: if a model's own description preserved the information it needed, re-reading that description could not reduce accuracy. Both point the same way, which is why we report the conclusion as approximately 85\% and not as a single exact figure.

\paragraph{Scope.} The probe covers two categories and two frontier systems, chosen because they are the settings where grounding is most clearly the binding constraint. We do not claim the 85\% figure transfers to OCR or medical items, where the failure structure differs; the handwriting category in particular has a documented cascade pattern in which perception and inference are entangled by construction (\Cref{app:failure}).

\section{Trace Alignment Protocol}
\label{app:align}

Each gold trace and each model response is segmented into steps by sentence and enumeration boundaries. Steps are embedded with \texttt{BAAI/bge-small-en-v1.5} using normalized embeddings, and for every gold step we take the maximum cosine similarity to any step of the model response. The item score is the mean of those per-step maxima, so the measure asks whether the model \emph{covered} the reference reasoning, not whether it matched it verbatim or matched its length. A response that reaches the same conclusion by a different valid route scores well as long as it touches the same intermediate facts.

For each model we then compute AUROC of this score against binary answer correctness over the ${\sim}1{,}000$ items. The length-controlled variant regresses response length out of the alignment score first and recomputes AUROC on the residual, which is what distinguishes a content effect from a verbosity artifact.

\Cref{tab:alignfull} reports per-model means and gaps for all 25 systems. The gap is positive in 25 of 25 cases, ranging from $+0.0177$ (GLM-4.1V-9B-Thinking) to $+0.0534$ (Qwen2.5-VL-7B-Instruct). Absolute similarities are high for every model and every outcome, since any on-topic response about the same image shares vocabulary with the reference. The gap, not the absolute value, therefore carries the signal.

\begin{table}[t]
\centering
\scriptsize
\caption{Trace alignment for all 25 evaluated systems, sorted by gap. ``Sim.'' is the mean over gold trace steps of the maximum cosine similarity to any step of the model response. The gap is positive for every variant: responses that reach the correct answer track the human reasoning trace more closely than responses that do not.}
\label{tab:alignfull}
\begin{tabular}{lccccc}
\toprule
\textbf{Model} & \textbf{$n$ correct} & \textbf{$n$ incorrect} & \textbf{Sim.\ (correct)} & \textbf{Sim.\ (incorrect)} & \textbf{Gap} \\
\midrule
Qwen2.5-VL-7B-Instruct & 274 & 719 & 0.8159 & 0.7625 & $+$0.0534 \\
OpenAI o3 & 570 & 408 & 0.8038 & 0.7515 & $+$0.0523 \\
VL-Rethinker & 291 & 701 & 0.8148 & 0.7657 & $+$0.0491 \\
GPT-5 & 633 & 361 & 0.8095 & 0.7644 & $+$0.0451 \\
Kimi-VL & 290 & 538 & 0.8368 & 0.7929 & $+$0.0439 \\
InternVL3.5-8B & 302 & 692 & 0.8138 & 0.7704 & $+$0.0434 \\
Visionary-R1 & 261 & 680 & 0.8023 & 0.7595 & $+$0.0428 \\
Claude Sonnet 4 & 396 & 541 & 0.8200 & 0.7787 & $+$0.0413 \\
Qwen2.5-VL-3B-Instruct & 213 & 695 & 0.7959 & 0.7569 & $+$0.0390 \\
Qwen3-VL-8B-Instruct & 400 & 555 & 0.8225 & 0.7840 & $+$0.0385 \\
Molmo-7B & 201 & 622 & 0.7936 & 0.7554 & $+$0.0382 \\
InternVL3-8B-Instruct & 267 & 694 & 0.8109 & 0.7732 & $+$0.0377 \\
Qwen3-VL-30B-A3B-Thinking & 367 & 506 & 0.8386 & 0.8018 & $+$0.0368 \\
InternVL3-2B-Instruct & 206 & 637 & 0.8028 & 0.7662 & $+$0.0366 \\
Qwen3-VL-30B-A3B-Instruct & 424 & 544 & 0.8220 & 0.7859 & $+$0.0361 \\
Gemma-3-12B-it & 346 & 650 & 0.8107 & 0.7747 & $+$0.0360 \\
Qwen3-VL-8B-Thinking & 360 & 584 & 0.8421 & 0.8070 & $+$0.0351 \\
Gemini 2.5 Pro & 658 & 336 & 0.8200 & 0.7856 & $+$0.0344 \\
Claude Opus 4.6 & 622 & 375 & 0.8279 & 0.7936 & $+$0.0343 \\
Gemma-3-4B-it & 290 & 706 & 0.8020 & 0.7680 & $+$0.0340 \\
Ovis2.5-9B & 361 & 625 & 0.8201 & 0.7886 & $+$0.0315 \\
Ovis2.5-2B & 284 & 706 & 0.8046 & 0.7738 & $+$0.0308 \\
Skywork-R1V2-38B & 366 & 583 & 0.8317 & 0.8034 & $+$0.0283 \\
Skywork-R1V3-38B & 385 & 603 & 0.8321 & 0.8085 & $+$0.0236 \\
GLM-4.1V-9B-Thinking & 237 & 734 & 0.7678 & 0.7501 & $+$0.0177 \\
\bottomrule
\end{tabular}
\end{table}

\section{Ranking Stability Protocol}
\label{app:stability}

The stability check in \Cref{sec:reliability} evaluates three frontier systems under four conditions on a 150-item stratified subset (30 items per category), for 1{,}799 valid runs with 0 errors. The conditions are \textbf{P0}, the default prompt used throughout the paper; \textbf{P1}, a terser instruction that drops the explicit formatting scaffold; \textbf{P2}, an explicit chain-of-thought instruction; and \textbf{P0+}, the default prompt at raised temperature. Every condition is a choice a competent practitioner might defensibly make. None is an adversarial perturbation.

Per-model accuracy is stable, deviating by at most 4.8 points from that model's own mean across conditions. The top-1 ranking is not: GPT-5 leads under P1 and Gemini 2.5 Pro under the other three, while Claude Opus 4.6 is last in all four. We therefore report that adjacent positions among closely matched systems are not identified by accuracy alone on this benchmark, and treat the grounding, faithfulness, and cost axes as the more informative comparison.

\section{Critic Validation Protocol}
\label{app:critic}

We sampled 200 model traces balanced across 5 categories, 8 open-weight models, and both outcomes (100 correct, 100 incorrect), and asked whether VisualPRM-8B assigns higher scores to correct reasoning. It separates the two by $+0.012$ overall (0.608 versus 0.596), and the gap is negative in multi-hop ($-0.037$) and coreference/occlusion ($-0.056$), the two categories that demand the most compositional and spatial grounding.

We report this because it bounds how \smarttwo{} should be read. \smart{} is critic-free and carries every headline result in this paper; \smarttwo{} is a critic-agnostic template whose validity is inherited from whichever judge is supplied, and with an off-the-shelf 8B critic on unseen domains that validity is currently weak. A stronger multimodal critic would improve \smarttwo{} without any change to the benchmark.

\section{Category Failure Modes}
\label{app:failure}

Expanding \Cref{sec:failure}, each category exhibits a distinct and deployment-relevant signature.

\paragraph{Visual coreference / occlusion.} Strong systems enumerate candidate entities, apply the spatial constraint to each, and eliminate. Weaker systems show a saliency bias, latching onto the most visually prominent entity regardless of the referring expression, and a spatial phrase inversion in which relations such as ``the man behind the woman in red'' have their figure and ground swapped.

\paragraph{Handwriting / OCR.} A single misrecognized character propagates through otherwise valid downstream arithmetic, producing a confidently wrong answer with an internally consistent chain. This is the clearest case where final-answer scoring and step-level scoring disagree about what went wrong, and where a correct-looking trace is actively misleading.

\paragraph{Navigation / map reasoning.} Landmark identification is generally reliable; path composition is not. Failures concentrate in off-by-one directional steps and left/right confusion at intersections. A second pattern is linguistic shortcutting, where the model infers cardinal orientation from street names rather than from the layout.

\paragraph{Multi-hop.} The characteristic pattern is first-hop success with second-hop fabrication: the model resolves the first clue and then introduces an unsupported constraint. Longer chains make this worse, not better, which is consistent with the reflective-token analysis in \Cref{sec:reflective}.

\paragraph{Scientific / medical.} Models frequently perceive the visual feature correctly and then hallucinate a plausible but incorrect causal attribution. This category has the highest rate of confidently wrong explanations and is the clearest safety concern for high-stakes deployment, since the stated reasoning reads as authoritative.

\section{Qualitative Examples}
\label{app:qualitative}

The figures below show one item per category, with the image, the answer options, the ground-truth rationale, and the full reasoning of four systems side by side with their generated token counts. They are included because the aggregate numbers in this paper describe distributions, and the mechanisms in \Cref{app:failure} are easier to assess against concrete items. Every panel is verbatim model output.

\begin{figure}[p]
\centering
\captionsetup{font=small}
\includegraphics[width=0.66\textwidth]{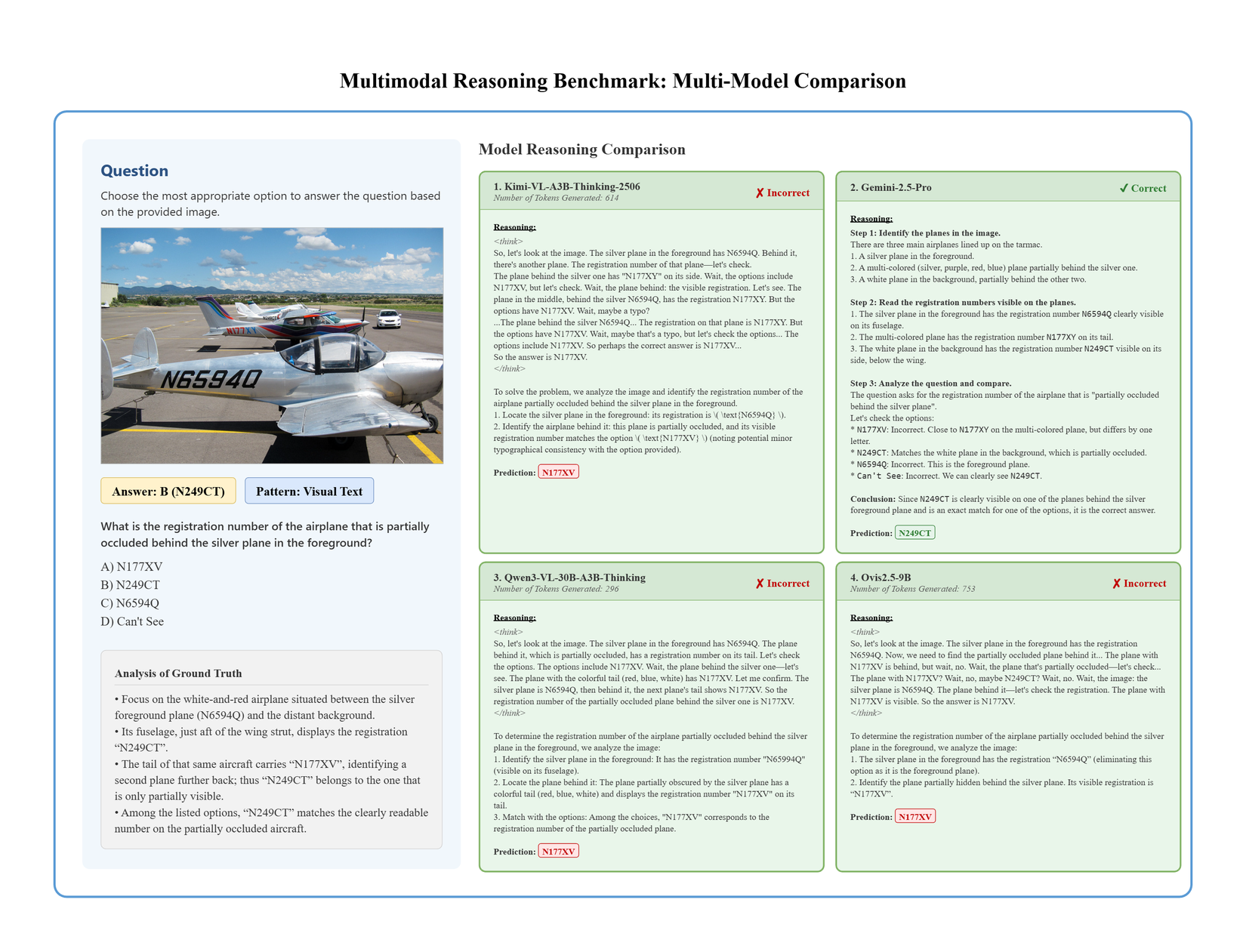}
\caption{\textbf{Visual coreference under occlusion.} The question asks for the registration of the aircraft \emph{partially occluded behind} the silver plane in the foreground. Gemini 2.5 Pro resolves it by enumerating the three visible aircraft, reading each registration, and matching against the referring expression. Kimi-VL, Qwen3-VL-30B-Thinking and Ovis2.5-9B all answer N177XV, the registration of the \emph{foreground} plane. This is the saliency bias described in \Cref{app:failure}: the referring expression is never applied, and the most prominent entity is returned instead. Token budgets of 614, 296 and 753 show this is not a matter of insufficient reasoning length.}
\label{fig:qual-coref}

\vspace{12pt}
\includegraphics[width=0.66\textwidth]{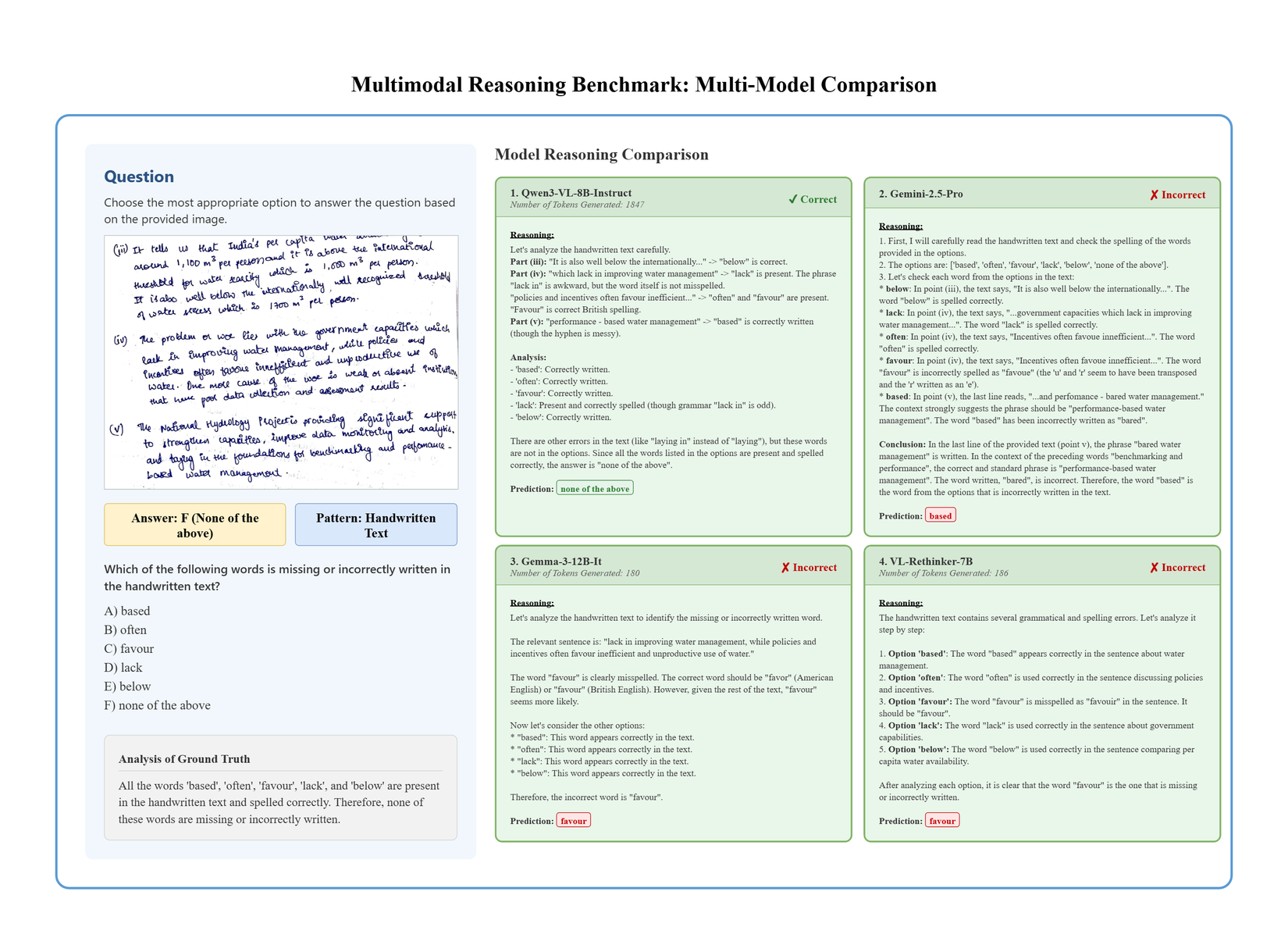}
\caption{\textbf{Handwriting / OCR reasoning.} All five candidate words appear correctly in the handwritten passage, so the answer is ``none of the above'', which requires transcribing every option correctly before concluding that nothing is missing. Qwen3-VL-8B-Instruct does so. Gemini 2.5 Pro misreads one word and then constructs an elaborate, internally consistent argument for ``based''; Gemma-3-12B and VL-Rethinker each settle on ``favour''. This is the cascade pattern: the reasoning is valid given the misperception, so the trace looks sound and the error is invisible without the ground truth.}
\label{fig:qual-ocr}
\end{figure}

\begin{figure}[p]
\centering
\captionsetup{font=small}
\includegraphics[width=0.66\textwidth]{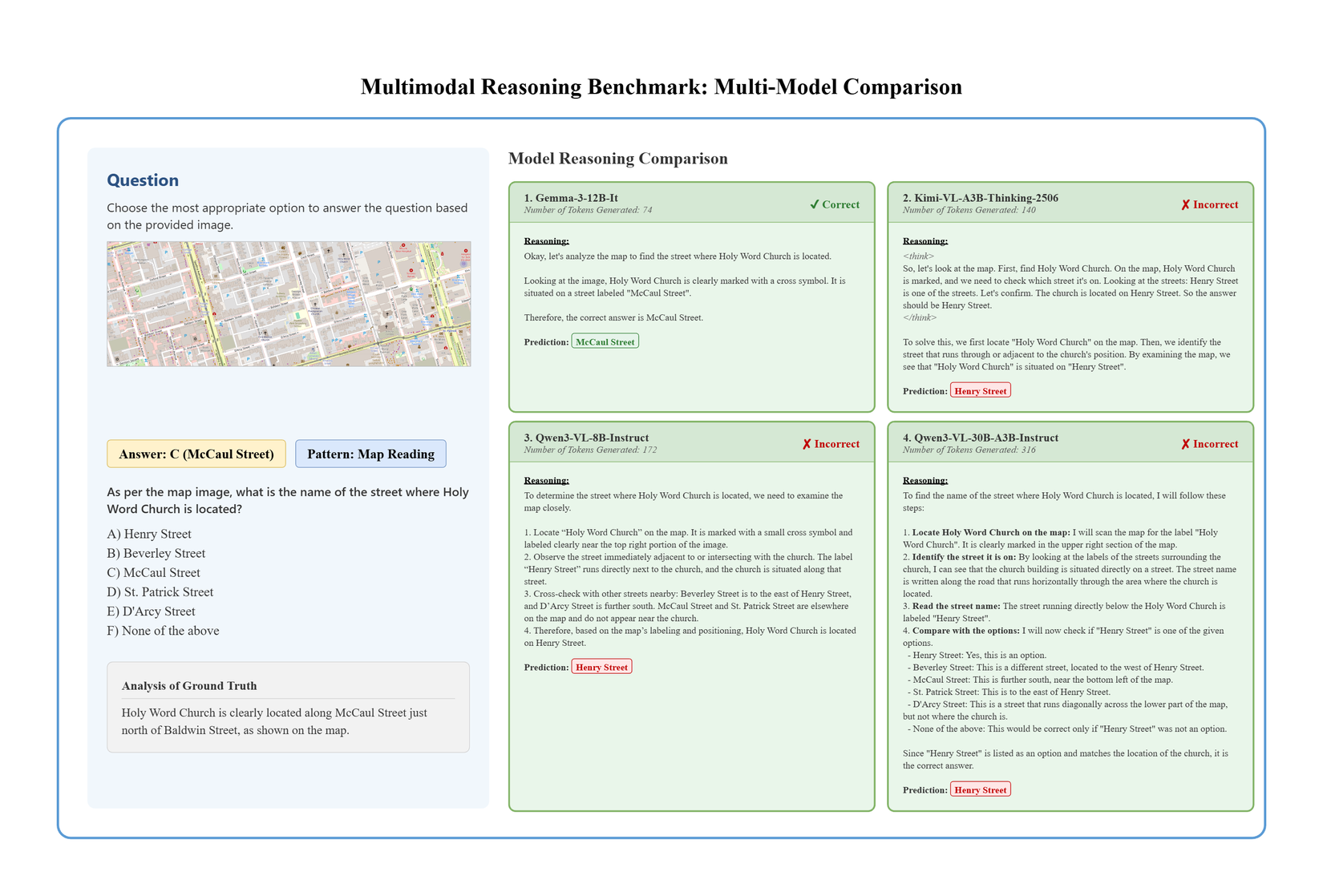}
\caption{\textbf{Navigation / map reading.} Gemma-3-12B-It reads the label adjacent to the church and answers correctly in \textbf{74 tokens}. Kimi-VL (140 tokens), Qwen3-VL-8B-Instruct (172) and Qwen3-VL-30B-Instruct (316) all produce longer, more structured deliberations and all answer ``Henry Street''. This item is a compact illustration of why we score cost alongside correctness: the cheapest response here is the only correct one, and a leaderboard ordered by accuracy alone would still rank three of these systems above Gemma overall.}
\label{fig:qual-nav}

\vspace{12pt}
\includegraphics[width=0.66\textwidth]{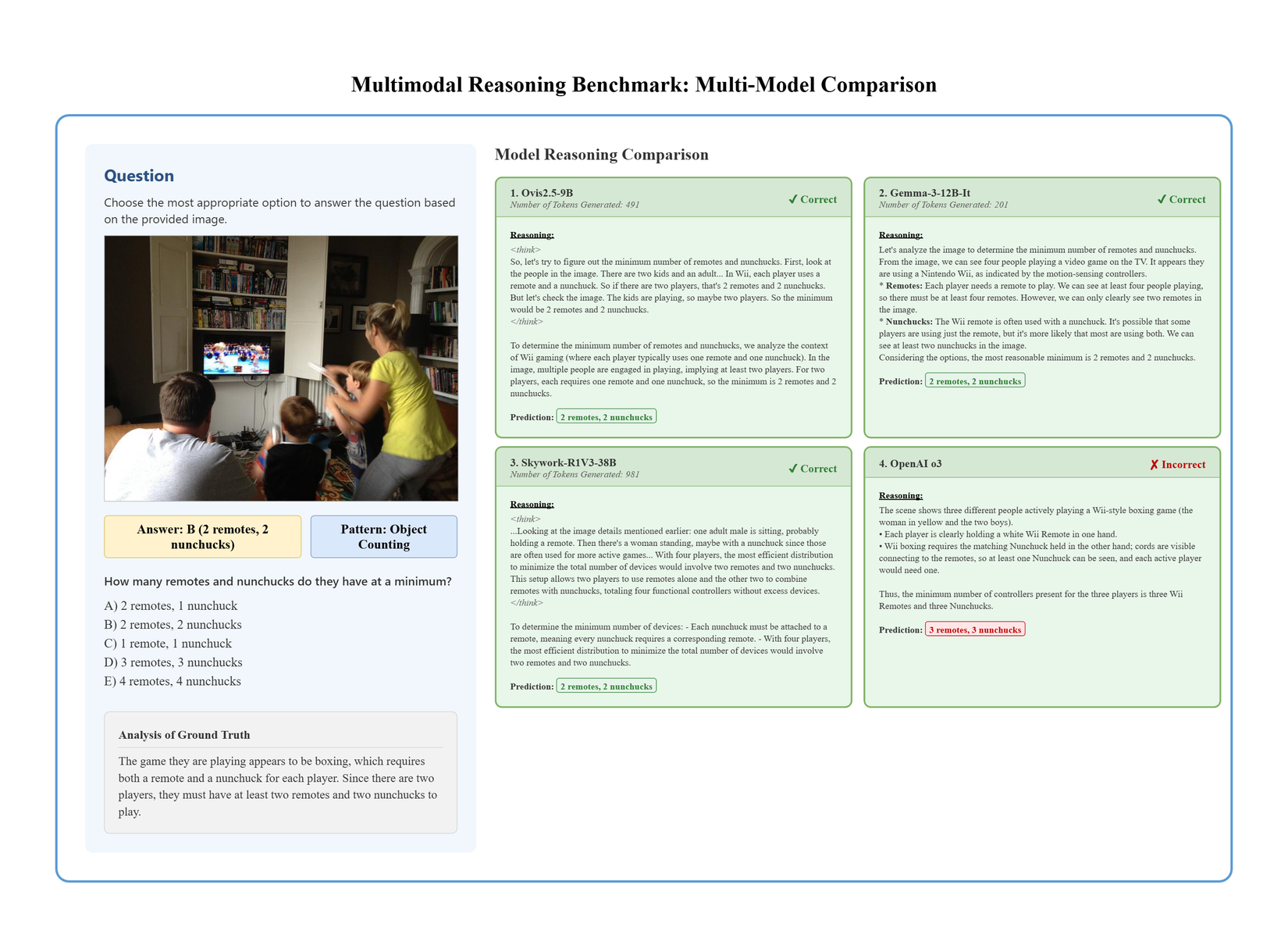}
\caption{\textbf{Multi-hop reasoning.} Answering requires chaining what the scene shows (people playing a Wii boxing game) to a fact about the game (each player needs a remote \emph{and} a nunchuck) to a count. Ovis2.5-9B, Gemma-3-12B and Skywork-R1V3-38B chain correctly. OpenAI o3 counts three players from the image and propagates that into three controllers, an error at the first hop that the subsequent reasoning faithfully preserves.}
\label{fig:qual-multihop}
\end{figure}

\begin{figure}[p]
\centering
\captionsetup{font=small}
\includegraphics[width=0.66\textwidth]{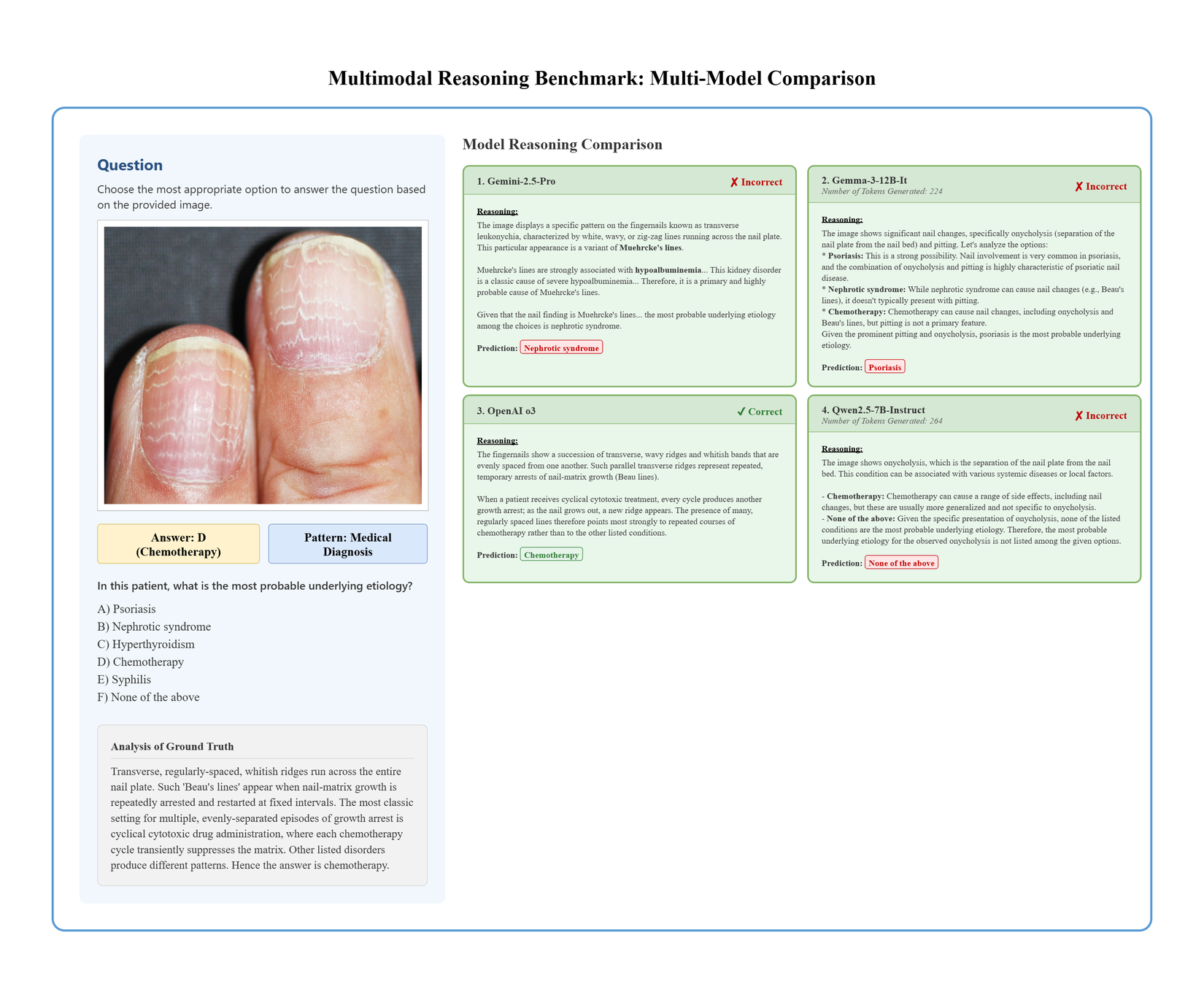}
\caption{\textbf{Scientific / medical reasoning.} The nail finding is Beau's lines, transverse ridges left by repeated arrest of nail-matrix growth, which point to cyclical chemotherapy. Only OpenAI o3 reaches this. Gemini identifies the pattern as Muehrcke's lines and reasons fluently to nephrotic syndrome; Gemma-3-12B reasons to psoriasis; Qwen2.5-VL-7B concludes none of the options apply. Each wrong answer arrives with a confident, clinically phrased causal chain. This is the category with the highest rate of confidently wrong explanations, and the reason we treat it as the clearest safety concern for deployment.}
\label{fig:qual-medical}

\vspace{12pt}
\includegraphics[width=0.66\textwidth]{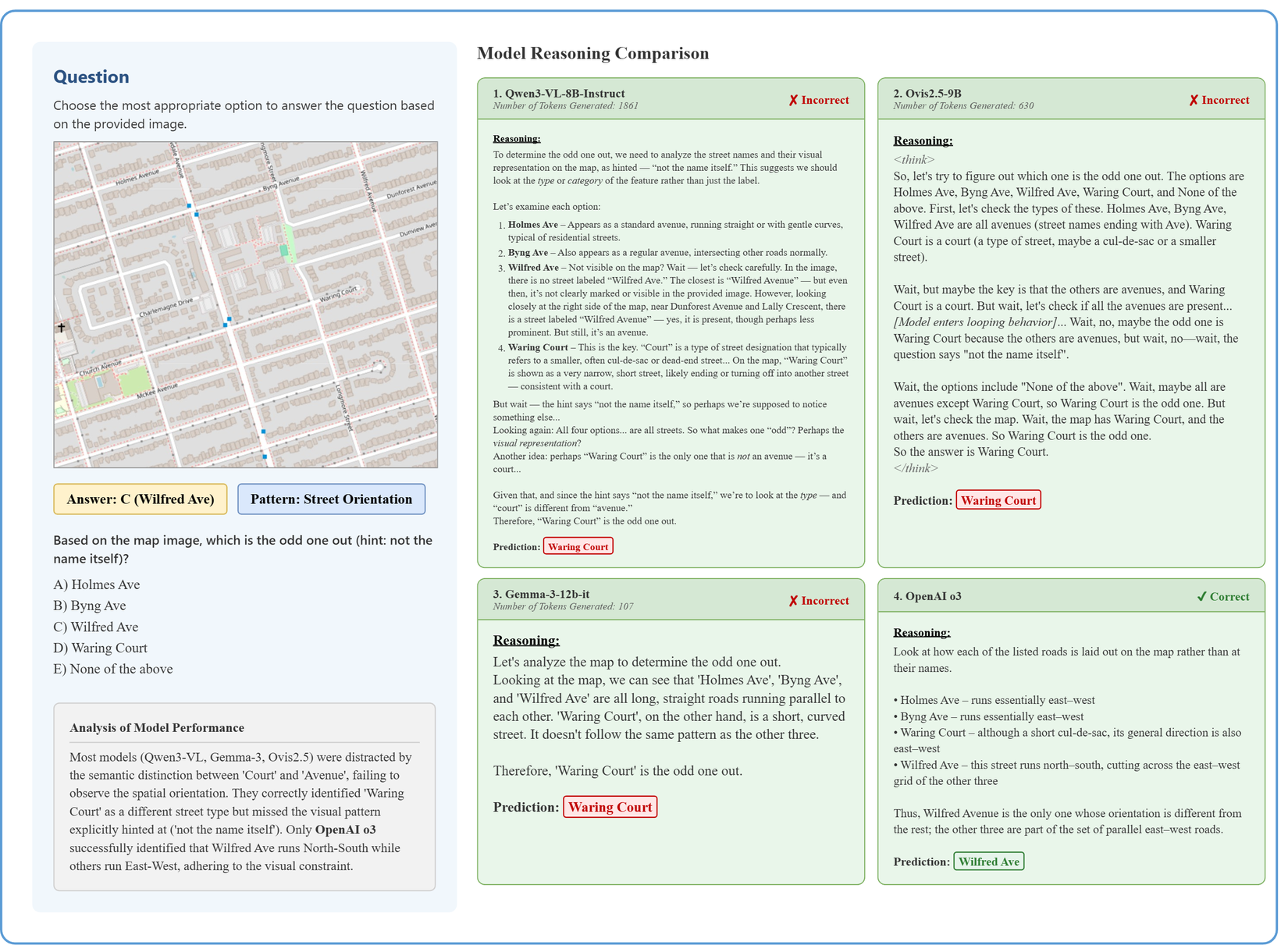}
\caption{\textbf{Navigation / spatial orientation.} The question asks which street is the odd one out and states explicitly that the criterion is not the name. The answer is Wilfred Ave, the only street running north--south while the others run east--west. Only OpenAI o3 checks orientation on the map. Qwen3-VL-8B-Instruct (1{,}861 tokens), Ovis2.5-9B and Gemma-3-12B all reason from the \emph{linguistic} distinction between ``Court'' and ``Avenue'' despite the hint ruling it out, and Ovis2.5-9B visibly loops before committing. This is the linguistic shortcutting failure, and it is the pattern behind the ${\sim}47\%$ navigation scores in \Cref{tab:fullresults}: the models read the map as text rather than as space.}
\label{fig:qual-orientation}
\end{figure}

\clearpage

\section{Benchmark Construction}
\label{app:construction}

\ours{} contains 1{,}000 items, 200 per category. Images are drawn from RefCOCO~\citep{refcoco} for coreference and multi-hop, schematic layouts derived from OpenStreetMap for navigation, handwriting scans and IAM for OCR, and MedXpertQA~\citep{medxpertqa} for the medical category.

\begin{figure}[t]
\centering
\includegraphics[width=\textwidth]{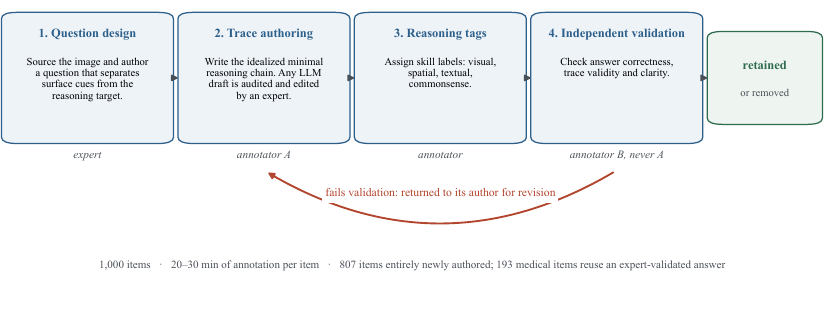}
\caption{The four-stage annotation loop. Duties are separated so that no annotator validates an item whose reasoning trace they wrote, and an item that fails Stage 4 returns to its author for revision instead of being silently accepted or discarded. Stage 2 produces the human reference length $L_i^{\mathrm{gt}}$ that \smart{} uses and the reference content that trace alignment uses.}
\label{fig:pipeline}
\end{figure}

\Cref{fig:pipeline} shows the annotation loop. Three annotators work across four stages, and the assignment of people to stages enforces a separation of duties: the annotator who validates an item is never the annotator who authored its trace. An item that fails validation does not simply get dropped. It returns to its author with the validator's objection, is revised, and is validated again; only items that pass an independent check are released, and items that cannot be repaired are removed. Each item takes 20--30 minutes of annotation effort across the four stages.

Two points about Stage 2 are worth stating plainly, because they affect how the reference traces should be read. First, where a language model was used to suggest an initial draft of a trace, that draft was mandatorily audited and edited by an expert annotator before acceptance; no trace enters the benchmark as raw model output. Second, the trace is an \emph{idealized minimal} chain, meaning the shortest logically complete path to the answer, which is what makes its length usable as an efficiency floor in \Cref{eq:core}.

Question, option, answer, and trace text is newly authored for 807 of 1{,}000 items (80.7\%); only source images are reused. The 193 medical items reuse an expert-validated image, question, and answer, with the trace authored against that verified answer. Internal 5-gram Jaccard overlap across questions is low (mean and median maximum pairwise similarity 0.21 and 0.11), so items are not near-duplicates of one another.

\paragraph{Contamination.} Because the source images are public, provenance documentation alone cannot establish that answers did not leak. We therefore ran a behavioral check, reported in \Cref{sec:reliability}: if exposure helped, the most-reused sources would be the easiest. They are the hardest. OpenStreetMap schematics sit at 46.8/47.8 for GPT-5 and Opus 4.6 and COCO/RefCOCO images at 60.6/61.3, while handwriting scans, the source least likely to have been memorized, are easiest at 82.1 and 83.6. Difficulty tracks the reasoning demanded rather than the familiarity of the pixels, which is what one expects when questions are written to ask about occluded content and multi-step spatial composition that no public caption contains.

\end{document}